\documentclass{article}

\usepackage{arxiv}
\usepackage[utf8]{inputenc} %
\usepackage[T1]{fontenc}    %
\usepackage[colorlinks,allcolors=black,pdftex]{hyperref}
\usepackage{url}            %
\usepackage{booktabs}       %
\usepackage{amsfonts}       %
\usepackage{nicefrac}       %
\usepackage{microtype}      %
\usepackage{xcolor}         %

\usepackage[american]{babel}

\usepackage{amssymb}
\usepackage{mathtools} %
\usepackage{xkeyval}
\usepackage{siunitx} %
\usepackage{booktabs,nicematrix} %
\usepackage{tikz} %
\usepackage{algpseudocode}
\usepackage{algorithm}

\let\vec\mathbf

\makeatletter
\patchcmd{\ALG@step}{\addtocounter{ALG@line}{1}}{\refstepcounter{ALG@line}}{}{}
\newcommand{\ALG@lineautorefname}{Line}
\makeatother

\usepackage{bbm}
\usepackage{cleveref}%
\usepackage{colortbl}
\usepackage{graphicx}
\usepackage{diagbox}
\usepackage{xspace}
\usepackage{caption}
\usepackage{subcaption}

\usepackage{multicol,multirow}
\usepackage{todonotes}

\usepackage{placeins} %
\usepackage{adjustbox}

\definecolor{RubineRed}{RGB}{237,1,125}
\definecolor{ForestGreen}{RGB}{0,155,85}

\usepackage{pifont}     %

\newcommand{\project}{BatMan\xspace}
\newcommand{\man}{Man\xspace}
\newcommand{\batman}{BatMan\xspace}

\newcommand{\data}[1]{\mathcal{D}_{\text{#1}}}
\newcommand{\dtest}{\data{test}}
\newcommand{\dsup}{\data{support}}
\newcommand{\dquery}{\data{query}}

\newcommand{\dtrain}{\data{train}}
\newcommand{\loss}[1]{\mathcal{L}_{#1}}

\title{\batman-CLR: Making Few-shots Meta-Learners Resilient Against Label Noise}
\author{
Jeroen M.,~Galjaard \\
{\tiny Delft University of Technology}\\
\texttt{j.m.galjaard@tudelft.nl}
\And
Robert,~Birke\\
{\tiny University of Turin}\\
\texttt{robert.birke@unito.it}
\\
\And
Juan F.,~P\'erez\\
{\tiny Universidad de los Andes}\\
\texttt{jf.perez33@uniandes.edu.co}\\
\And
Lydia Y.,~Chen\\
{\tiny Delft University of Technology}\\
\texttt{y.chen-10@tudelft.nl}
}
\hypersetup{
pdftitle={BatMan-CLR: Making Few-Shot Meta-Learners Resilient Against Label Noise},
pdfsubject={cs.ML},
pdfauthor={Jeroen M.~Galjaard, Juan F.~Perez, Robert~Birke, Lydia Y.~Chen},
pdfkeywords={meta-learning,few-shot,label noise,robust learning},
}

\begin{document}
\maketitle

\begin{abstract}
The negative impact of label noise is well studied in classical supervised learning yet remains an open research question in meta-learning. 
Meta-learners aim to adapt to unseen learning tasks by learning a good initial model in meta-training and consecutively fine-tuning it according to new tasks during meta-testing.
In this paper, we present the first extensive analysis of the impact of varying levels of label noise on the performance of state-of-the-art meta-learners, specifically gradient-based $N$-way $K$-shot learners.
We show that the accuracy of Reptile, iMAML, and foMAML drops by up to 42\% on the Omniglot and CifarFS datasets when meta-training is affected by label noise. 
To strengthen the resilience against label noise, we propose two sampling techniques, namely manifold (\man)
and batch manifold (\batman), which transform the noisy supervised learners into semi-supervised ones to increase the utility of noisy labels.
We first construct manifold samples of $N$-way $2$-contrastive-shot tasks through augmentation, learning the embedding via a contrastive loss in meta-training, and then perform classification through zeroing on the embedding in meta-testing. 
We show that our approach can effectively mitigate the impact of meta-training label noise.
Even with 60\% wrong labels \batman and \man can limit the meta-testing accuracy drop to ${2.5}$, ${9.4}$, ${1.1}$ percent points, respectively, with existing meta-learners across the Omniglot, CifarFS, and MiniImagenet datasets. 
\end{abstract}

\section{Introduction}\label{sec:intro}
Few-Shot Learning (FSL) poses the problem where learners need to quickly adapt to new unseen tasks by using a low number of samples.
Meta-learning~\cite{Schmidhuber1987,Finn2017} emerged as a promising solution to this problem.
Like humans, meta-learners learn the information at a higher abstraction or meta-level, providing the inductive bias to adapt to new tasks quickly.
Among existing meta-learners, gradient-based few-shot learners, e.g., iMAML~\cite{Rajeswaran2019} and foMAML(+ZO)~\cite{Finn2017,Kao2021}, have been shown 
effective to solve $N$-way $K$-shot $(N, K)$ problems, that need to learn $N$ classes given only $K$ samples each.
Such few-shot learners are composed of two stages, meta-training and meta-testing, each with their own labeled support and query data sets.
Meta-training learns a meta-model using two sequential optimization loops. The inner-loop adapts the model to a specific task via supervised learning on the support set. The outer-loop updates the meta-model based on the adapted task-specific model and query set.
Using a similar structure, meta-testing verifies how well the meta-model performs on new tasks. First, it uses supervised learning to adapt the meta-model to an unseen task given by a test support set. Then it computes the learner's accuracy on the testing query set comparing predicted and given labels.  
Class labels are thus crucial in both meta-training and meta-testing.

Label noise is more the norm than a rarity and can significantly degrade the performance of supervised learners~\cite{noisesurvery22}.
Prior studies address label noise mainly in classical supervised learners. In this context, samples hold labels different from the underlying ground truth.  
In the context of FSL, label noise means that a shot (example) may not correspond to the way (class) it was provided with. This yields a degenerate $N$-way $K$-shot problem where ways become indistinguishable since they contain shots of the same ground truth.
Such noise may appear in the meta-training and meta-testing support and meta-training query set.

Given the importance of labels in meta-training and meta-testing, only a few studies~\cite{Mazumder2021RNNP,Liang2022,Lu2021} address the challenge of noisy labels in FSL and %
only at the level of the meta-testing support set.
As the number of samples per class is very limited, e.g., five to ten shots, the task adaptation step can be over-parameterized by label noise and lead to significant degradation. 
Unfortunately, label noise can appear at all the support and query sets of meta-testing and meta-training, and little is known on its impact and resolution. 
Moreover, the existing meta-learners that account for label noise still require clean data to learn a meta-objective. 
\cite{Killamsetty2022}
distills the impact of label noise by adding an additional meta-objective on clean validation data.
\cite{Yao} proposes an adaptive task-aware scheduling (ATS) to learn to filter out \emph{noisy tasks} but assumes a static set of tasks, of which only a fraction has corrupted support sets.

In this paper, we first answer whether state-of-the-art FSL methods are resilient to label noise. 
We consider label noise present in the query and support sets during meta-training. 
We empirically show that Reptile~\cite{Nichol2018}, Eigen-Reptile~\cite{pmlr-v162-chen22aa}, iMAML~\cite{Rajeswaran2019}, and foMAML+ZO~\cite{Kao2021} are significantly affected by label noise in meta-training, overfitting to noise and degrading the efficacy of any randomly initialized models. 
To address the noisy labels in meta-training, we propose \batman, which turns any supervised few-shot learner into a semi-supervised one by a novel batch manifold sampling and contrastive learning.
We learn the embedding in meta-training and then apply a zeroing strategy in meta-testing.
Specifically, we turn a noisy $N$-way $K$-shot problem into a self-cleansed $N$-way $2$-\emph{contrastive} shot problem.
We first augment the original shots and construct contrastive pairs, ensuring the shots are from the same class. 
We then sample such pairs from the $N$ ways, termed manifold (\man) samples.
To lower the probability of getting noisy $N$-ways, i.e., overlapping classes, we draw a batch of such \man samples, termed \batman sampling. 
Combining this approach with the Decoupled Contrastive Loss (DCL)~\cite{Yeh2022}, we can effectively learn the embedding of the initial model, which can then be adapted in meta-testing to a new $N$-way $K$-shot task.

The specific contributions of this paper are:
\begin{enumerate}
  \setlength{\itemsep}{-2.5pt}
    \item A first-of-its-kind study on the impact of label noise in meta-training for gradient-based meta-learners. 
    \item A generic and self-cleansing framework, \batman-CLR, that turns meta-learners into semi-supervised ones by (batched) manifold sampling $N$-way $2$-contrastive shots.
    \item Extensive evaluation on four meta-learners, Reptile, EigenReptile, iMAML, and foMAML, shows nearly no performance degradation under the presence of up to 60\% label noise.
\end{enumerate}

\noindent\begin{figure*}[!ht]
    \centering
    \begin{subfigure}{0.4\linewidth}
        \centering
        \includegraphics[height=12em,clip,trim=1cm 1.9cm 0.5cm 0.1cm
    ]{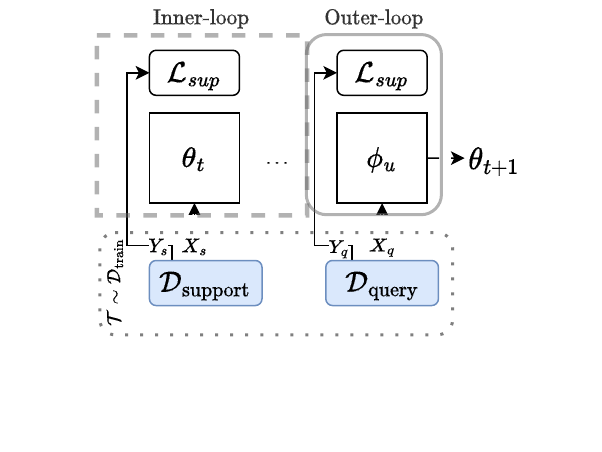}
        \caption{Meta-training}\label{fig:pipeline-training}
    \end{subfigure}
    ~
    \begin{subfigure}{0.55\linewidth}
    \centering
    \includegraphics[height=12em,clip,trim=1cm 1.9cm 0.5cm 0.1cm
    ]{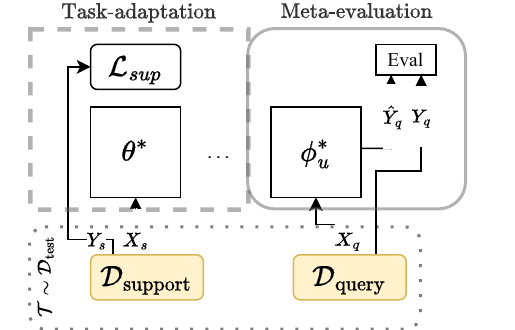}
    \caption{Meta-testing.}\label{fig:pipeline-testing}
    \centering
    \end{subfigure}
    \caption{ 
   High-level overview of meta-learning (left) and meta-testing (right) with gradient-based meta-Learners.
   During each meta-epoch's $t$ inner-loop, the meta-model $\theta_t$ is adapted to a task $\mathcal{T}$ in few steps $u$
   to a task-specific model $\phi_u$, leveraging $(X_s, Y_s) \in \dsup$.
   To find the updated meta-model $\theta_{t+1}$, the learner relates $\phi_u$'s loss on $(X_q, Y_q) \in \dquery$ back to $\theta_t$. 
   Meta-testing evaluates the capability of trained meta-model $\theta^{*}$ to adapt to a task specific model $\phi_u^{*}$ via $\dsup$. The performance of $\phi_u^{*}$ is then evaluated on $\dquery$. %
   At each meta-epoch, the learner selects a task $\mathcal{T}$ from $\dtrain$ and $\dtest$ to sample $(\dsup, \dquery)$ for meta-training and meta-testing.
}\label{fig:pipeline}
\end{figure*}

\section{Preliminary and Related work}\label{sec:related-work}

\textbf{Preliminary on FSL.}
FSL considers the setting where a learned model must adapt to new settings leveraging only a few samples. 
In this setting, meta-learning has emerged as a promising research direction.
Gradient-based meta-learners aim to find a meta-model with parameters $\theta$ capable of quickly adapting into a task-specific parameterization $\phi$. 
We consider the $N$-way $K$-shot classification problem, which consists of a family of tasks, made up of $N$ classes, each with $K$ samples, termed $(N,K)$-FSL tasks.   
We focus on gradient-based meta-learners which aim to find $\theta$ iteratively using a two-step meta-training algorithm (see \autoref{fig:pipeline-training}).
At the beginning of each meta-epoch, the learner selects a task $\mathcal{T}$ %
and samples two task-specific sets, support $\dsup$ and query $\dquery$, from the training data $\dtrain$. 
More formally a task $\mathcal{T}$ is a tuple of support and query data $(\dsup,\dquery) = \mathcal{T}$, defined as sets of inputs $X$ and targets (labels) $Y$:
\begin{align*}
\dsup  = \bigcup_{i=1}^{N} \{({X}^{i}_j, Y^{i}_{j})\} 
_{j=1}^{K},\, \dquery = \bigcup_{i=1}^{N} \{({X}^{i}_j, Y^{i}_{j})\} 
_{j=1}^{Q}.
\end{align*}
Next, step one transforms $\theta_t$ using the features and labels $(X_s, Y_s) \in \dsup$ and a supervised loss function $\loss{sup}$ in task-specific parameters $\phi$.
Then, step two uses $\phi$, the data $(X_q, Y_q) \in \dquery$, and $\loss{sup}$ to obtain $\theta_{t+1}$ for the next iteration.
Note that while the support set $\dsup$ is used during an inner-loop to train for a specific task, the query set $\dquery$ is used in an outer-loop to learn the meta-model.
During meta-training, the support set for a single task is built by randomly selecting $N$ classes from the training set $\dtrain$, each with $K$ samples. 
For the query set, we select $Q$ additional samples for the 
same $N$ selected classes.

Meta-testing aims to evaluate the ability of the trained meta-model $\theta^*$ to adapt to new tasks (see \autoref{fig:pipeline-testing})
Analogous to meta-training, first we select a testing task and sample $\dsup$ and $\dquery$ from the test data $\dtest$. Next, an adaptation step uses $\dsup$ to transform the generic meta-model with parameters $\theta^*$ into the task-specific model with parameters $\phi^*$. Finally, $\phi^*$ is tested on the query set $(X_q,Y_q) \in \dquery$ by comparing its predictions $\hat{Y}_q$ against known labels $Y_q$.

Examples of gradient-based meta-learners include MAML~\cite{Finn2017}, iMAML~\cite{Rajeswaran2019}, foMAML(+ZO)~\cite{Kao2021},  Reptile~\cite{Nichol2018} and Eigen-Reptile~\cite{pmlr-v162-chen22aa}.
MAML updates the meta-model via gradient descent through gradient descent~\cite{Finn2017}.
As this operation is both compute and memory intensive~\cite{Finn2017,Rajeswaran2019,MuhammadAbdullahJamal2021} many works proposed approximations. %
iMAML~\cite{Rajeswaran2019} and foMAML (with the zero-ing trick from~\cite{Kao2021}) approximate the meta-gradient w.r.t. $\theta_t$ %
by leveraging the first-order gradient on $\dquery$ w.r.t. $\phi_{u}$. 
This drops the need for gradient descent through gradient descent.

foMAML assumes that the higher-order components of the meta-gradient can be ignored altogether, whereas iMAML enforces that it can %
calculate the meta-gradient using more adaptation steps %
and weight regularization.
Reptile~\cite{Nichol2018} %
drops $\dquery$ during meta-training by estimating the meta-gradient by stepping towards $\phi_{u}$ to find $\theta_{t+1}$. %
Eigen-Reptile~\cite{pmlr-v162-chen22aa} builds on this by decomposing the inner-optimization path $[\theta_t, \phi_1,\ldots,\phi_{u}]$ and stepping towards the direction with the largest variance. %

\textbf{Label noise in FSL.}
Label noise poses a major challenge to meta-learners, especially in the absence of clean data that can be used as ground truth during the training phase.  
Although a large collection of work exists on robust supervised learning~\cite{Wang2020,Li2020,Yu2020}, these are not directly applicable to meta-learners due to the limited number of samples available during each adaptation process. 
Recognizing the presence of label noise, related studies~\cite{Yao,Lu2021,Killamsetty2022,Liang2022,Mazumder2021RNNP,Liang2022} mainly focus on distilling the label noise appearing in \emph{meta-testing} by explicitly studying the noise patterns~\cite{Yao,Liang2022,Lu2021}, using soft-relabeling~\cite{Mazumder2021RNNP} through clustering or re-weighting suspicious samples~\cite{Killamsetty2022} based on additional %
\emph{clean} data.
To our best knowledge, Eigen-Reptile~\cite{pmlr-v162-chen22aa} is the only study that addresses noisy \emph{training data} in FSL by updating the inner-loop only along the direction of the highest variance. 
However, such an approach %
lacks generalization to other state-of-the-art meta-learners.

\begin{table}
\newcolumntype{H}{>{\setbox0=\hbox\bgroup}c<{\egroup}@{}}
\centering
\caption{Meta-test accuracies on clean meta-test data, following training on varying levels of meta-train noise. 
Each learner was validated with 2048 (5, 5)-FSL meta-test tasks with transductive inference, using a query size of 15. `+ZO' indicates trained with Zero-ing trick~\cite{Kao2021}.}\label{tab:baseline-noisy-meta-test}
\begin{subtable}[H]{0.6\linewidth}
\centering
\begin{tabular}{lccc}
\toprule
Algorithm & $\epsilon=0.0$       & $\epsilon=0.3$           & $\epsilon=0.6$       \\ \midrule 
Reptile       & 65.5{\tiny±0.241} & 58.6{\tiny±0.258} & 51.8{\tiny±0.253} \\
foMAML+ZO     & 69.5{\tiny±0.255} & 65.2{\tiny±0.265} & 40.3{\tiny±0.207} \\
iMAML         & 64.0{\tiny±0.242} & 55.9{\tiny±0.257} & 46.3{\tiny±0.243} \\ \midrule
Eigen-Reptile & 65.3{\tiny±0.243} & 58.1{\tiny±0.272} & 52.7{\tiny±0.241} \\ \bottomrule
\end{tabular}
\caption{CifarFS results.} %
\end{subtable}
~
\begin{subtable}[H]{0.35\linewidth}
\centering
\begin{tabular}{HHccc}
\toprule
Alg. &  & $\epsilon=0.0$       & $\epsilon=0.3$           & $\epsilon=0.6$       \\ \midrule 
Reptile     & NaN  & 92.5{\tiny±0.125} & 79.7{\tiny±0.214} & 71.5{\tiny±0.240} \\
foMAML+ZO   & NaN  & 99.3{\tiny±0.037} & 97.7{\tiny±0.067} & 90.3{\tiny±0.148} \\
iMAML       & NaN  & 96.9{\tiny±0.110} & 91.0{\tiny±0.176} & 82.6{\tiny±0.192} \\ \midrule
ER          & NaN  & 93.6{\tiny±0.122} & 83.6{\tiny±0.189} & 73.4{\tiny±0.235} \\ \bottomrule
\end{tabular}
\caption{Omniglot results.}
\end{subtable}
\end{table}

\textbf{The impact of label noise.} 
Here we motivate the need of noise resilience in few shot learning via an empirical study.
We investigate the effect of label noise on two representative datasets, CifarFS~\cite{Bertinetto} and Omniglot~\cite{Lake2011}, in a (5, 5)-FSL setting with a query set of 15 samples per class. 
The details of the datasets and experiments can be found in \autoref{sec:setup}. 
We re-implement four different meta-learners: Eigen-Reptile (ER)~\cite{pmlr-v162-chen22aa}, Reptile~\cite{Nichol2018}, first order MAML with Zero Out (foMAML+ZO)~\cite{Kao2021}, and implicit MAML (iMAML)~\cite{Rajeswaran2019}.  We train each learner using hyper-parameters comparable to the ones in the corresponding paper.
We consider a symmetric label noise setting, where each label of class $i$ has $\epsilon$ probability to be corrupted with a random label $j \neq i$  with uniform probability across all other classes.

\autoref{tab:baseline-noisy-meta-test} shows the meta-test accuracy obtained by the different meta-learners under varying degrees of corrupted training labels, $\epsilon = [0.0, 0.3, 0.6]$. Note that $\epsilon=0.0$ means no noise, i.e., all clean labels. Reported results are the average across three runs with $95$\% confidence intervals. Both datasets and all meta-learners clearly show a significant performance degradation as the noise ratio increases. 
With the CifarFS dataset, accuracy drops across all meta-learners on average by $10.1$\% and $27.4$\% under 30\% and 60\% corrupted labels, respectively. 
The Omniglot dataset shows similar trends but more limited in amplitude, with $8.1$\% and $17.0$\% average degradation. This is due to the fact that the Omniglot dataset is easier to learn. Indeed, all meta-learners obtain an accuracy score above 90\% under zero noise. Interestingly, although ER is the sole meta-learner that explicitly aims to counter noise, it is not always the most robust one. Under moderate noise, i.e.~30\%, foMAML+ZO is the least affected with accuracy drops of $6.2$\% and $1.6$\% for CifarFS and Omniglot, respectively. Only under heavy noise, i.e.~60\%, on the CifarFS dataset ER is the least affected learner (accuracy drop of $19.3$\%) and able to beat the others by a small 0.9 percentage points higher accuracy margin. 
More in general, Reptile, ER and iMAML show a higher but almost linear impact of noise, while foMAML+ZO degrades less with 30\% noise but gets much %
worse under 60\% noise. Overall, the results underline the need for better noise resilience across all meta-learners.

\section{Proposed method}
\begin{figure}[t]
    \centering
    \includegraphics[width=1.0\linewidth,clip, trim=0cm 0cm 2.8cm 0cm]{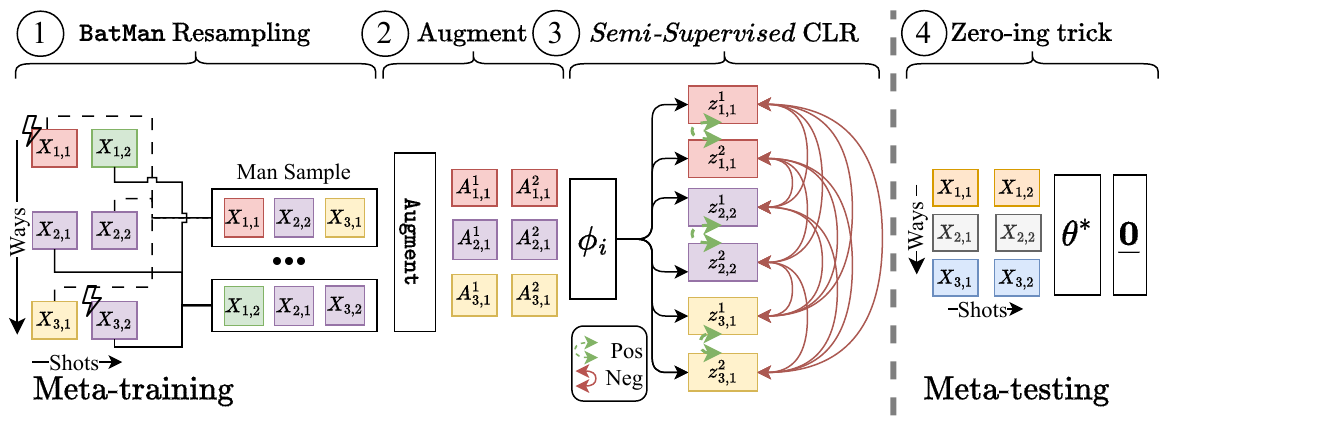}
    \caption{
    Overview of meta-training and testing with \batman-CLR on a noisy 3-way 2-shot $(3, 2)$-Few-Shot Learning (FSL) task. Colors represent label \emph{ground truth}. Lightning bolts indicate corrupted samples (i.e., way 1 shot 1, and way 3 shot 2). Steps:
    1) \man sampling (\autoref{code:mansampling}) creates a batch of (3, 1)-FSL manifold samples; 2) \texttt{Augment} creates semi-supervised (3, 2)-FSL sub-tasks via independent random augmentations; 3) a contrastive loss jointly optimizes all sub-tasks via (Pos)itive and (Neg)ative pairs; 4) we meta-test in a supervised way the trained meta-model $\theta^*$ by appending a zero-ed out linear layer.
    }\label{fig:method}
\end{figure}

The core challenge of dealing with noise in meta-(or few shot) learning is that labels lose meaning misguiding standard supervised approaches with wrongly labeled samples. 
This challenge is amplified in the FSL setting as the limited number of samples (shots) in each class (way) makes it harder to isolate the noise from the signal in each class. 
In other words, the few clean samples---those corresponding to the original not-corrupted class%
---may not be enough to appropriately guide the gradient descent algorithm in the correct direction of the class label. %
Thus, our approach aims at building clean ways and shots---such that each `way' becomes more likely to have samples all with the same ground truth. %
Specifically, the proposed \man sampling 
uses data augmentation to create clean shots,  
which guarantees that the underlying ground truth label is the same,
rather than leveraging \emph{other} shots of the same `way'.
We further introduce batches, with \batman sampling, to increase the likelihood of observing all $N$ classes in a single inner-loop step.
In this section, we go over the three main steps of our proposed method, namely, 
i) re-sampling, either with \man or \batman, 
ii) `semi'-supervised meta-training with a contrastive loss, and 
iii) classification of new tasks (meta-testing) leveraging a zeroing trick. 
Together these steps can be incorporated into \emph{existing meta-learning algorithms} to achieve %
label noise robustness.
\autoref{sec:results} provides results for both \man and \batman re-sampling. %
\autoref{fig:method} provides a graphical depiction of our proposed method applied on a noisy (3, 2)-FSL task.

\noindent

\begin{minipage}[h]{\linewidth}
\begin{minipage}[t]{0.49\linewidth}
\begin{algorithm}[H]
\begin{algorithmic}[1]
\Require $\texttt{Augment}$ feature augment function.
\Function{ManSampling}{$\mathcal{D}, N$}
\State $M = \{\}$
\For{$j \in [1, \ldots, N]$}
    \State $X \gets rand\left((X,y) \in \data{} : y = j\right)$
    \State $A^1 \gets \texttt{Augment}(X)$
    \State $A^2 \gets \texttt{Augment}(X)$
    \State $M \gets M \cup \{(A^1, j), (A^2, j)\}$
\EndFor
\State\Return $M$
\EndFunction
\end{algorithmic}
\caption{Pseudocode for \man Sampling. 
\batman is achieved through multiple \man samples.}\label{code:mansampling}
\end{algorithm}
\end{minipage}
~
\begin{minipage}[t]{0.50\linewidth}
\centering
    \begin{algorithm}[H]
\centering
    \caption{
    General %
    structure with
BatMan-CLR for MAML style learners.}\label{algorithm1} %
    \begin{algorithmic}[1]
    \Function{\texttt{BatMan}-CLR}{$\phi$, $\dsup$, $\dquery$, $\alpha$, $N$, $u$}
        \For{$B_i \in [\texttt{BatMan}(\dsup, N)]_{i=1}^{u}$}\label{line:sampler}
            \State $\phi \gets \phi - \frac{\alpha}{v} \nabla{}_{\phi}  
            \sum\limits_{M_j \in B_i }
            \loss{con}(\phi( M_j)))$
        \EndFor
        \State $B_{\text{query}} \gets \texttt{BatMan}(\dquery, N)$
        \State $\mathcal{L}_{outer} \gets \sum\limits_{M_j \in B_{\text{query}}}
            \loss{con}(\phi( M_j)))$
        \State\Return $\phi, \mathcal{L}_{outer}$
    \EndFunction
    \end{algorithmic}

\end{algorithm}
\end{minipage}
\end{minipage}

\textbf{Manifold and \batman re-sampling.} 
We start with an $(N, K)$-FSL problem with noisy labels and observe that we can re-frame it as an $(N, 2)$-FSL problem by sampling $1$ data point from each way
and creating one more sample for each class employing augmentations.
Thus we end up with two samples with the same label for each way.
Since we sample $N$ observations from all $N\times K$ in $\dquery$, we end up with \emph{up to $N$ actual classes}, as some classes may end up contributing more than one observation to a sub-sampled task. %
In this manner, many potential $(N, 2)$ sub-tasks can be created, each corresponding to a different task `manifold'~\cite{Nichol2018} due to noisy labels. 
We coin this sampling approach Manifold (\man) sampling, described as pseudo-code in~\autoref{code:mansampling}. 
Note that to avoid introducing a bias towards the original sample, for each shot we use \texttt{Augment} to draw two random augmentations.   
As a simple extension of \man sampling, we propose a Batched Manifold (\batman) sampling, where several \man samples are batched together to employ more samples in a single step.
By grouping a batch of \man samples, we can jointly optimize multiple ``sub-problems''.
Additionally, this increases the likelihood of considering \emph{all} query classes together in the calculation of meta-gradients. 

This process is illustrated in~\autoref{fig:method}, with a 3-way 2-shot FSL setting, where each row index represents a way and each column index a shot.
The classes are illustrated with different colors, where initially there are 2 shots of each of the green, purple, and yellow classes. 
Noise has caused the class of the first shot in the green class to have a red class label, and the second shot of the yellow class to have a purple class label. 
A pool of \man samples is generated by sampling 1 shot from each of the 3 classes and augmenting it twice to have an $(N,2)$ FSL task. 
A batch of such tasks is generated to perform a meta-training step through gradient descent.

In the presence of noise, the process of cleaning up the classes is necessary as samples from two apparently different classes may actually come from the same class. 
In the simplest case with 2 classes and a probability $p = 1 - \epsilon$ that a sample has a ground truth label, the classes resulting from the \man sampling process belong to different classes with probability $p^2+(1-p)^2$, as either both lack noise or both are noisy. 
On the contrary, the 2 samples selected actually belong to the same class with probability $2p(1-p)$. 
In general, with $N$ classes there are $N!$ combinations in which the \man samples actually correspond to the $N$ \emph{different} classes while there are $N^N$ combinations to select the $N$ samples, with replacement. 
Let us consider the case of symmetric noise where a sample has a ground truth label $i$ with probability $p$ and that the sample is mislabeled as a different class $j\neq i$ with probability $(1-p)/(K-1)$. 
The probability of obtaining a clean selection of classes can then be posed as the probability of obtaining one of the $N!$ combinations in which this occurs. 
To represent each possible selection we employ permutation matrices. Let $P^i_N$ be the $N\times N$ $i$th permutation matrix out of the $N!$ such matrices. For instance, in the $N=3$ case, we have 6 different permutation matrices, i.e., 
\[
\begin{bmatrix}
1 &0 &0\\
0 &1 &0\\
0 &0 &1\\
\end{bmatrix},\ \begin{bmatrix}
1 &0 &0\\
0 &0 &1\\
0 &1 &0\\
\end{bmatrix},\ \begin{bmatrix}
0 &1 &0\\
1 &0 &0\\
0 &0 &1\\
\end{bmatrix},\ \begin{bmatrix}
0 &1 &0\\
0 &0 &1\\
1 &0 &0\\
\end{bmatrix},\ \begin{bmatrix}
0 &0 &1\\
1 &0 &0\\
0 &1 &0\\
\end{bmatrix},\ \begin{bmatrix}
0 &0 &1\\
0 &1 &0\\
1 &0 &0\\
\end{bmatrix}. 
\]
Let us also define the $N\times N$ matrix $Q$ with entries $q_{ij}$ such that $q_{ii}=p$ and $q_{ij}=(1-p)/(N-1)$ for $i\neq j$.
The probability of obtaining one of these valid permutations under \man sampling can thus be obtained as the trace of the matrix $P^i_NQ$. 
As a result, the probability of obtaining a clean selection of ways can be expressed as 
\[
\sum_{i=1}^{N!} \text{trace}(P^i_NQ),
\]
considering all possible valid selections of samples that lead to a set with $N$ different classes. %
As this probability becomes smaller with increasing label noise, \batman sampling helps by introducing additional samples that increase the likelihood of observing all $N$ classes in a single meta-epoch.

\textbf{Semi-supervised meta-training with contrastive loss.} 
Although re-sampling allows for likely valid $(N, 2)$ sub-tasks, \emph{which classes} they contain remains unknown.
As such these sub-tasks can be considered as \emph{semi-supervised}, as there are now at most $N$ classes. %
To aid this, we incorporate a contrastive loss to allow for joint optimization of semantically misaligned sub-tasks. 
Note that from the sampled augmentations obtained in step 1 of~\autoref{fig:method}, we artificially build positive and negative pairs. 
These positive and negative pairs can be optimized under a contrastive learning strategy. 
We use the Decoupled Contrastive Loss (DCL)~\cite{Yeh2022}, which is particularly well-suited for small data sets, although alternative contrastive losses can easily replace it. 
Once samples are augmented and \batman sampling applied, the batches are used in the meta-training step where the embeddings $z$ are computed and contrasted using a contrastive loss.

\textbf{Classification of new tasks}. The meta-model $\theta^*$ trained using the contrastive loss produces embeddings instead of classes as output. 
To solve this, we append the meta-model with a fully connected layer $C_\vec{0} = (\vec{W},\underline{\vec{b}})$ with $\vec{W} = \vec{0}$ and $\underline{\vec{b}} = \underline{\vec{0}}$. 
This approach decouples the embedding learning from the classification task.
Similar to~\cite{Kao2021}, this allows us to treat the model as a semi-supervised meta-learned backbone. %
The resulting meta-model $\theta^{*\prime} = C_{\vec{0}} \circ \theta^{*} $ can then be treated as a supervised learner utilizing the cross entropy (CE) classification loss.
We found that applying the Zeroing Out trick on the \emph{classification} layer significantly impacts the learner's performance because it allows leveraging the optimized embedding from the pre-trained meta-model $\theta^{*}$.
This is because the stochastic gradient descent will directly use the embeddings as activations without the noise introduced by randomly initialized weights.

\section{Evaluation  Results}\label{sec:results}
In this section, we present the effectiveness of \batman-CLR in enhancing the noise resilience for state-of-the-art meta-learners, namely Reptile, iMAML, and foMAML+ZO, under the presence of different noise levels. We further include Eigen-Reptile, a noise-aware FSL as an additional baseline.

\subsection{Setup}
\label{sec:setup}
We consider three data sets in a (5, 5)-FSL setting: Omniglot, Cifar Few-Shot (CifarFS), and miniImagenet. 
To emulate training label noise, for each dataset, we add symmetric random noise to $\dtrain$ with 30\% and 60\% corrupted labels. 
For instance, for experiments with 60\% noise, we randomly select 60\% samples of each class in $\dtrain$, and assign them to a different class \emph{within the same split}.
Reported meta-test results are on \emph{clean} data, to evaluate the learners under a base-case scenario. 
The support set is of size 5 (15), and query set 15 (Reptile learners), following~\cite{Finn2017,Nichol2018}. 
All experiments ran on machines with 128 GB RAM, 2x AMD EPYC 7282 16-Core CPUs, and a 16 GB Nvidia A4000 GPU. We used cross-entropy loss for supervised meta-training and meta-testing.

We incorporate \man and \batman sampling into 
Reptile, Eigen-Reptile, iMAML and foMAML with Zeroing-trick.
Each learner uses a ConvNet-4 architecture with 64 filters and a linear layer with output dimension $\mathbb{R}^{128}$.
On Omniglot, the number of filters was increased to 128.
We use weight decay centered around $\theta_t$~\cite{Rajeswaran2019} on iMAML and foMAML+ZO resets its final layer to zero at the beginning of each inner-loop. 
The learners were trained with the original papers' hyper-parameters, except for the following changes. 
Eigen-Reptile and Reptile run with $7$ inner-loop steps, iMAML with $12$ ($16$ for Omniglot), and foMAML with $5$.
iMAML's proximal decay was set to $0.5$ ($2.0$ for Omniglot).
Each learner was meta-tested after 5K, 15K (10K), 15K (10K), training outer-loop steps (iMAML) respectively for Omniglot, CifarFS and MiniImagenet.
On CifarFS and MiniImagenet we use the augmentations proposed in~\cite{Bachman}.
For Omniglot, %
we %
follow the normal augmentation scheme in~\cite{Boutin2022}, applying one of random crop, affine transform, or perspective transform. 
During meta-testing the task-specific model is fine-tuned for 10 steps on Omniglot and CifarFS, and 20 steps on MiniImagenet.

When applying \batman-CLR on the meta-learners, we keep the same model sizes with the addition of a larger decision head: $\mathbb{R}^{128}$ rather than $\mathbb{R}^{5}$.
The batch size of \batman is set to 5 for all inner-loop adaptations, thereby resulting in mini batching for (Eigen) Reptile, and 15 for the meta-gradient calculation of iMAML and foMAML. 
For each support sample, 5 augmentations are created, whereas each query sample is augmented twice, allowing the inner-loop to sample more diverse tasks.
The final reported testing accuracy is averaged over 3 test runs, using 2048 tasks sampled from $\dtest$.

\subsection{Testing accuracy}

\begin{table*}[t]
\newcolumntype{H}{>{\setbox0=\hbox\bgroup}c<{\egroup}@{}}
\newcommand\RotText[1]{\fontsize{9}{9}\selectfont \rotatebox[origin=c]{90}{{\centering#1}}}
\newcolumntype{g}{>{\columncolor{gray!20}}c}
\caption{(Meta-test results  with $95$th Confidence Interval ($\overline{acc}${\tiny ± CI95}) of Meta-Pretrained models (5, 5)-FSL with different noise levels $\epsilon$ during training.} %
\label{tab:batman-clr}
\begin{subtable}[T]{0.55\linewidth}
\centering
\begin{tabular}[t]{lHlccc}
\toprule
{Algorithm}                         & views & Sampler & $\epsilon$=0.0   & $\epsilon$=0.3   & $\epsilon$=0.6  \\ \midrule
\multirow{2}{*}{Reptile}    
                                                          & 5                                                & BatMan                       & 66.5{\tiny±0.168} & 65.0{\tiny±0.170} & 64.1{\tiny±0.170} \\
                                                          &                                                  & Man                          & 61.8{\tiny±0.176} & 62.0{\tiny±0.175} & 61.4{\tiny±0.173} \\ \midrule
\multirow{2}{*}{Eigen Reptile}                        & 5                                                    & BatMan                       & 66.3{\tiny±0.171} & 64.4{\tiny±0.170} & 63.8{\tiny±0.176} \\
                                                          &                                                  & Man                          & 61.7{\tiny±0.170} & 55.8{\tiny±0.378} & 52.3{\tiny±0.365} \\ \midrule
\multirow{2}{*}{foMAML}                     & 5                                                              & BatMan                       & 66.6{\tiny±0.167} & 65.2{\tiny±0.169} & 64.8{\tiny±0.164} \\
                                                          &                                                  & Man                          & 66.2{\tiny±0.164} & 64.8{\tiny±0.165} & 64.0{\tiny±0.165} \\ \midrule
\multirow{2}{*}{iMAML}                      & 5                                                              & BatMan                       & 64.2{\tiny±0.185} & 62.7{\tiny±0.250} & 62.9{\tiny±0.203} \\
                                                          &                                                  & Man                          & 62.8{\tiny±0.172} & 62.6{\tiny±0.168} & 61.7{\tiny±0.169} \\ \bottomrule
\end{tabular}
\caption{CifarFS results}
\end{subtable}
\hfill
\begin{subtable}[t]{0.35\linewidth}
\centering%
\begin{tabular}[t]{HHHccc}
\toprule
{Alg.}                         & views & Sampler & $\epsilon$=0.0   & $\epsilon$=0.3   & $\epsilon$=0.6  \\ \midrule
\multirow{2}{*}{\RotText{Reptile}}     
                                                      & 5                                                    & BatMan   & 97.9{\tiny±0.070} & 97.3{\tiny±0.078} & 96.2{\tiny±0.100} \\
                                                          &                                                  & Man      & 97.8{\tiny±0.068} & 97.8{\tiny±0.068} & 97.7{\tiny±0.070} \\ \midrule
\multirow{2}{*}{\RotText{IER}}                         & 5                                                   & BatMan   & 92.6{\tiny±0.128} & 93.0{\tiny±0.119} & 93.2{\tiny±0.117} \\
                                                          &                                                  & Man      & 93.7{\tiny±0.116} & 93.9{\tiny±0.114} & 94.0{\tiny±0.111} \\ \midrule
\multirow{3}{*}{\RotText{fM}}                     & 5                                                        & BatMan   & 98.2{\tiny±0.066} & 98.2{\tiny±0.063} & 98.0{\tiny±0.067} \\
                                                          &                                                  & Man      & 98.1{\tiny±0.062} & 98.1{\tiny±0.062} & 98.1{\tiny±0.061} \\ \midrule
\multirow{3}{*}{\RotText{iM}}                      & 5                                                       & BatMan   & 97.5{\tiny±0.078} & 98.1{\tiny±0.069} & 98.3{\tiny±0.063} \\
                                                          &                                                  & Man      & 97.8{\tiny±0.076} & 98.2{\tiny±0.074} & 98.2{\tiny±0.064} \\ \bottomrule
\end{tabular}
\caption{Omniglot results.}
\end{subtable}
\end{table*}

The results of \batman-CLR on noisy CifarFS and Omniglot are summarized in~\autoref{tab:batman-clr}, and on MiniImagenet in~\autoref{tab:miniimagenet-batman}. We report testing accuracy with both \man and \batman for CifarFS and Omniglot.
Prior to analyzing the results, we underline how all learners face significant degradation under label noise, as shown in~\autoref{sec:related-work}.

\textbf{Omniglot and CifarFS}. \batman-CLR clearly strengthens the resilience of all learners on all three datasets, with only a marginal decrease in testing accuracy under label noise.  
On Omniglot (see~\autoref{tab:batman-clr}, when encountering the label noise in meta-training, all learners can still learn effective initial models for task adaptation, reaching an accuracy of around 96\%.
In fact, all learners are able to display a performance under label noise similar to that without noise. 
On CifarFS, we observe similar results. Most learners reach a test accuracy between 62-64\%, except when applying \man sampling on Eigen-Reptile. 
These results strongly validate the effectiveness of \batman, which self-cleanses the `shots' by creating contrastive pairs and `ways' in batched \man samples. 

In terms of comparison between \batman and \man, there is a visible advantage in using \batman, especially on the more difficult CifarFS. 
This suggests that taking steps with more information, as in \batman, provides greater benefits than taking a larger number of simpler steps, as in \man.  
Zooming into the performance of different learners on CifarFS, the difference in testing accuracy between \man and \batman on CifarFS is smaller with foMAML and iMAML, compared to Reptile and Eigen-Reptile.
This can be explained by the fact that in our experiments, the MAML style learners use \batman to calculate the meta-gradient, resulting in more informative meta-updates.
Reptile learners do not calculate their meta-gradients using query data but directly using the inner-optimization \emph{direction}.

An observation worth mentioning is that Eigen-Reptile paired with \man, deteriorates under noise on CifarFS.
We speculate that this is due to the fact that in Eigen-Reptile the meta-gradient approximation is performed by selecting the optimization direction with the highest variance. 
However, a high level of noise introduces a high variance into the optimization directions, making it harder to select an appropriate direction even with the use of \man. 
By employing the less noisy \batman estimation strategy, the learner is able to better select an optimization direction and achieves a performance comparable to Reptile.

\begin{table}[t]
\newcolumntype{H}{>{\setbox0=\hbox\bgroup}c<{\egroup}@{}}
\centering
\caption{Meta-test accuracy with $95$th Confidence Intervals ($\overline{acc}${\tiny ± CI95}) \batman on (5, 5)-FSL MiniImagenet under varying label noise levels ($\epsilon$). 
White and gray columns correspond to supervised and \project-CLR results, respectively. Supervised foMAML uses the zeroing trick~\cite{Kao2021} (+ZO).}\label{tab:miniimagenet-batman}
\newcolumntype{a}{>{\columncolor[gray]{0.8}}c}
\begin{NiceTabular}[colortbl-like]{@{}l ca  ca  ca }
\toprule
Algorithm   & \multicolumn{2}{c}{$\epsilon$=0.0}     &   \multicolumn{2}{c}{$\epsilon$=0.3}    &   \multicolumn{2}{c}{$\epsilon$=0.6}   \\ \midrule
Reptile  & 54.2{\tiny ±0.206}       &  53.2{\tiny ±0.163}      & 27.8{\tiny ±0.141}       &  52.8{\tiny ±0.145}      & 24.6{\tiny ±0.143}       &  52.1{\tiny ±0.147}      \\
Eigen Reptile  & 58.7±{\tiny 0.250}       &  50.9{\tiny ±0.146}      & 44.8{\tiny ±0.200}       &  50.5{\tiny ±0.148}      & 24.8{\tiny ±0.140}       &  50.5{\tiny ±0.144}      \\
foMAML(+ZO)  & 52.2{\tiny ±0.217}       &  51.5{\tiny ±0.216}      & 37.1{\tiny ±0.179}       &  51.2{\tiny ±0.218}      & 28.2{\tiny ±0.145}       &  51.5{\tiny ±0.217}      \\
iMAML   & 53.9{\tiny ±0.215}       &  50.4{\tiny ±0.218}      & 45.5{\tiny ±0.211}       &  50.5{\tiny ±0.221}      & 20.0{\tiny ±0.076}       &  50.4{\tiny ±0.212}      \\ \bottomrule
\end{NiceTabular}
\end{table}

\textbf{MiniImagenet}. We further evaluate \batman-CLR on MiniImagenet, a more difficult dataset consisting of more diverse classes and larger inputs. \autoref{tab:miniimagenet-batman} summarizes the results.
Under \batman-CLR, the testing accuracy of most meta-learners has only limited drops, i.e., ranging between 0.5 and 1\%.

\subsection{Ablations}
First, we consider the impact of supervised task generation by training meta-learners in a self-supervised learning (SSL) setting.
Similar to UMTRA~\cite{Khodadadeh2019} and CACTUS~\cite{Hsu2019}, we construct $(5, 5/15)$-FSL tasks (MAML/Reptile) by drawing $5$ random images from $\dtrain$.
To construct the required shots, $K + Q$ augmentations are created and split in a support and query set, such that $|\dsup|=K$ and $|\dquery|=Q$.
\autoref{tab:umtra-robin} provides the results on Omniglot and CifarFS in the rows indicated with SSL.
We use the same hyper-parameters and loss function as for \batman-CLR, with the meta-batch size increased to $25$ (from $5$), so that learners see a comparable number of unique classes per meta-epoch.

Although this approach shows similar performance to \batman-CLR on Omniglot, on CifarFS there is a considerable gap of 11-13.8 percent points compared to \batman-CLR (gray columns).
This indicates that \batman-CLR profits from seeing \emph{more} unique classes and samples during each inner-loop. %

\begin{table}[!t]
\centering
\caption{Meta-test accuracy with $95$th confidence intervals ($\overline{acc}${\tiny ± CI95}), meta-trained on (5, 5)-FSL tasks with varying degrees of meta-training label noise ($\epsilon$) on Omniglot (a-b) and CifarFS (c-d), and different Inner/Outer-loop samplers: Random Manifold (R), and \batman (B).
SSL represents a self-supervised meta-trained model on unsupervised few-shot data.
\autoref{tab:batman-clr} results marked in gray.
}\label{tab:umtra-robin}
\newcommand\RotText[1]{\fontsize{9}{9}\selectfont
  \rotatebox[origin=c]{90}{\parbox{3em}{%
\centering#1}}}
\newcommand{\cc}[0]{\cellcolor[gray]{0.8}}
\newcolumntype{a}{>{\columncolor[gray]{0.8}}c}
\newcolumntype{H}{>{\setbox0=\hbox\bgroup}c<{\egroup}@{}}
\begin{subtable}[H]{0.55\linewidth}
\centering
\begin{tabular}{@{}clcccc@{}}
\toprule
\phantomsubcaption{(a)}\label{subtab:omniglot-maml}         & $\epsilon$  & B/B               & B/R               & R/B               & R/R               \\ \midrule
\multirow{4}{*}{\RotText{foMAML}}  & SSL                  & \multicolumn{4}{c}{94.7{\tiny±0.119}}                                         \\ 
& $0.0$ & \cc 98.2{\tiny±0.066} & 94.1{\tiny±0.123} & 98.3{\tiny±0.048} & 94.5{\tiny±0.120} \\

             & $0.3$                & \cc 98.2{\tiny±0.063} & 96.9{\tiny±0.089} & 98.4{\tiny±0.047} & 96.7{\tiny±0.091} \\
             & $0.6$                &\cc98.0{\tiny±0.067} & 97.9{\tiny±0.072} & 98.4{\tiny±0.047} & 98.0{\tiny±0.073} \\ %
            \midrule
\multirow{4}{*}{\RotText{iMAML}}   & SSL                  & \multicolumn{4}{c}{97.0{\tiny±0.090}}                                         \\ 
  & $0.0$ &\cc 97.5{\tiny±0.078} & 94.9{\tiny±0.114} & 98.1{\tiny±0.071} & 94.5{\tiny±0.119} \\
             & $0.3$                &\cc 98.1{\tiny±0.069} & 96.5{\tiny±0.097} & 98.3{\tiny±0.066} & 96.7{\tiny±0.091} \\
             & $0.6$                &\cc 98.3{\tiny±0.063} & 97.8{\tiny±0.075} & 98.2{\tiny±0.068} & 97.9{\tiny±0.073} \\ %
            \bottomrule
\end{tabular}
\end{subtable}
\quad
\begin{subtable}[H]{0.3\linewidth}
\centering
\begin{tabular}{@{}clcc@{}}
\toprule
\phantomsubcaption{(b)}\label{subtab:omniglot-reptile}  & $\epsilon$         & B                & R               \\ \midrule
\multirow{4}{*}{\RotText{Reptile}} & SSL                                      & \multicolumn{2}{c}{96.0{\tiny±0.098}}                                         \\
& $0.0$                                & \cc 97.9{\tiny±0.070}  & 65.3{\tiny±0.287} \\
             & $0.3$                                    & \cc 97.3{\tiny±0.078}  & 69.5{\tiny±0.271} \\
             & $0.6$                                    & \cc 96.2{\tiny±0.100}  & 73.1{\tiny±0.272} \\  %
              \midrule
\multirow{4}{*}{\RotText{Eigen Reptile}}  & SSL                                      & \multicolumn{2}{c}{95.1{\tiny±0.134}}                                         \\
& $0.0$                                    & \cc 92.6{\tiny±0.128}  & 82.5{\tiny±0.207} \\
             & $0.3$                                    & \cc 93.0{\tiny±0.119}  & 71.9{\tiny±0.270} \\
             & $0.6$                                    & \cc 93.2{\tiny±0.117}  & 73.5{\tiny±0.259} \\ %
             \bottomrule
\end{tabular}
\end{subtable}

\begin{subtable}[H]{0.55\linewidth}
\centering
\begin{tabular}{@{}clcccc@{}}
\toprule
\phantomsubcaption{(c)}\label{subtab:cifarfs-maml}               & {$\epsilon$} & B/B               & B/R               & R/B               & R/R               \\ \midrule
\multirow{4}{*}{\RotText{foMAML}} & SSL            & \multicolumn{4}{c}{52.8{\tiny±0.158}}                                         \\ 
& $0.0$ &\cc 66.6{\tiny±0.167} & 58.4{\tiny±0.236} & 62.0{\tiny±0.234} & 58.0{\tiny±0.236} \\
             & $0.3$          & \cc65.2{\tiny±0.169} & 59.7{\tiny±0.232} & 61.8{\tiny±0.236} & 59.7{\tiny±0.233} \\
             & $0.6$          &\cc 64.8{\tiny±0.164} & 60.5{\tiny±0.235} & 61.1{\tiny±0.237} & 60.5{\tiny±0.243} \\ %
            \midrule
\multirow{4}{*}{\RotText{iMAML}}   & SSL            & \multicolumn{4}{c}{54.5{\tiny±0.236}}                                         \\ 
& $0.0$ & \cc 64.2{\tiny±0.185} & 58.0{\tiny±0.238} & 60.9{\tiny±0.296} & 56.9{\tiny±0.288} \\
             & $0.3$          &\cc 62.7{\tiny±0.250} & 59.2{\tiny±0.240} & 60.3{\tiny±0.234} & 58.3{\tiny±0.287} \\
             & $0.6$          & \cc62.9{\tiny±0.203} & 59.7{\tiny±0.238} & 60.3{\tiny±0.286} & 60.0{\tiny±0.289} \\ %
           \bottomrule
\end{tabular}
\end{subtable}
\quad
\begin{subtable}[H]{0.3\linewidth}
    \begin{tabular}{@{}clcc@{}}
\toprule
\phantomsubcaption{(d)}\label{subtab:cifarfs-reptile}       & $\epsilon$         & B                & R               \\ \midrule
\multirow{4}{*}{\RotText{Reptile}}   & SSL            & \multicolumn{2}{c}{55.0{\tiny±0.167}}   \\ 
& $0.0$          & \cc66.5{\tiny±0.168}  & 53.9{\tiny±0.220} \\
             & $0.3$          &\cc 65.0{\tiny±0.170}  & 55.3{\tiny±0.219} \\
             & $0.6$          &\cc 64.1{\tiny±0.170}  & 56.4{\tiny±0.221} \\ %
            \midrule
\multirow{4}{*}{\RotText{Eigen Reptile}} & SSL            & \multicolumn{2}{c}{54.5{\tiny±0.164}}                 \\ 
& $0.0$          & \cc 66.3{\tiny±0.171}  & 57.2{\tiny±0.272} \\
             & $0.3$          & \cc 64.4{\tiny±0.170}  & 57.8{\tiny±0.227} \\
             & $0.6$          &\cc  63.8{\tiny±0.176}  & 58.8{\tiny±0.227} \\ %
             \bottomrule
    \end{tabular}
\end{subtable}
\end{table}
 
Second, we replace \batman with a random manifold sampler (Rand) to investigate the impact of the \batman sampling strategy.
\autoref{tab:umtra-robin} shows the results.
We pair Rand with \batman in different configurations for the inner and outer-loop, again evaluated on Omniglot and CifarFS.
Reptile and Eigen-Reptile only use a support set during meta-training, so we only replace their inner-sampling strategy.
We keep the same hyper-parameters as used in the corresponding \batman-CLR setting.

In general, the learners trained with random sampling in the outer-loop show an increased accuracy as the noise level increases.
Learners show an increase in accuracy of around 2-8\% and 2-3\% on Omniglot and CifarFS, comparing noise-less with $\epsilon=0.6$, whereas \batman-CLR sees a slight drop while staying ahead of Rand across the board.
This shows that \batman has the capability to self-clean.
An interesting exception is Omniglot combined with Eigen-Reptile when increasing the noise level to $0.3$ from $0$ (\autoref{subtab:omniglot-reptile}).
This is expected as higher noise levels increase the \emph{expected} number of unique ground-truth classes in a task, thereby yielding fewer false negatives in each Rand manifold during contrastive learning. 
Replacing only the inner or outer-loop sampler for iMAML and foMAML with Rand, we see that the contribution of the inner-loop is less significant than the outer-loop. 
This shows that \batman-CLR is also an effective strategy when replacing only the outer-loop (R/B).

\section{Conclusion}

Motivated by the ubiquitous presence of label noise, we empirically unveil the impact of label noise on existing few-shot meta-learners, with a particular focus on noise in meta-training. As the number of shots per class is low, the label noise can be exceedingly detrimental to meta-learners and highly challenging to address. 
To enhance the resilience against label noise for few-shots learners, we propose \batman---a generic approach that turns supervised few-shot tasks into semi-supervised ones. 
\batman is capable of self-cleansing noisy $N$-way $K$ shots instances by (i) batch manifold sampling that re-constructs $N$-way $2$-contrastive-shots via augmentation and (ii) introducing the DCL~\cite{Yeh2022} contrastive loss. 
Our results on three datasets, Omniglot, CifarFS, and MiniImagenet, show that \batman can maintain the effectiveness of few-shot learners independent of label noise levels, i.e., reserving %
almost 30\% accuracy degradation.

As future work, we aim to explore further (label noisy) meta-testing paired with \batman-CLR and adding class awareness~\cite{Khodadadeh2019,Shirekar2022}.
Further exploring the utility under the meta-testing setting of \batman-CLR would be valuable to this work.
Other loss functions can be explored, such as ProtoCLR~\cite{Medina}.
Lastly, we consider limited types of label noise during meta-training, leaving for future work the extension to noise such as out-of-domain noise, asymmetric noise, or task-level corruption~\cite{Yao}, as well as taking classes' semantics into account during noise generation.

\bibliographystyle{splncs04}
\bibliography{bibliography}
\end{document}